%% file: main.tex
\documentclass[letterpaper, conference]{IEEEtran}
\IEEEoverridecommandlockouts
\usepackage{cite}
\usepackage{amsmath,amssymb,amsfonts}
\usepackage{algorithmic}
\usepackage{graphicx}
\usepackage{textcomp}
\usepackage{xcolor}
\usepackage{svg}
\usepackage[nolist]{acronym} 
\usepackage{verbatim}
\usepackage[hidelinks]{hyperref}
\def\BibTeX{{\rm B\kern-.05em{\sc i\kern-.025em b}\kern-.08em
    T\kern-.1667em\lower.7ex\hbox{E}\kern-.125emX}}

\begin{document}

\title{Exploring Silent Data Corruption as a Reliability Challenge in LLM Training}

\newif\ifanonymous
\anonymousfalse

\ifanonymous
  \author{Anonymous Authors}
\else
    \author{
        \IEEEauthorblockN{Anton Altenbernd}
        \IEEEauthorblockA{
            Technische Universität Berlin\\
            Berlin, Germany \\
            a.altenbernd@tu-berlin.de
        }
        \and
        \IEEEauthorblockN{Philipp Wiesner}
        \IEEEauthorblockA{
            Technische Universität Berlin\\
            Berlin, Germany \\
            wiesner@tu-berlin.de
        }
        \and
        \IEEEauthorblockN{Odej Kao}
        \IEEEauthorblockA{
            Technische Universität Berlin\\
            Berlin, Germany \\
            odej.kao@tu-berlin.de
        }
    }
\fi

\maketitle

\input{content}

\bibliographystyle{IEEEtran}
\bibliography{refs}

\end{document}

%% file: content.tex
\begin{acronym}
    \acro{AI}{Artificial Intelligence}
    \acro{SDC}{Silent Data Corruption}
    \acro{LLM}{Large Language Model}
    \acro{ECC}{Error-Correction Code}
    \acro{RDMA}{Remote Direct Memory Access}
    \acro{MSB}{Most Significant Bit}
    \acro{DNN}{Deep Neural Networks}
    \acro{ABFT}{Algorithm-based Fault-Tolerance}
    \acro{CDF}{Cumulative Distribution Function}
    \acro{SM}{Streaming Multiprocessor}
    \acro{HMMA}{Half-Precision Matrix-Multiply Accumulate}
    \acro{GEMM}{General Matrix Multiply}
    \acro{RMS}{Root Mean Square}
    \acro{SASS}{Streaming ASSembler}
\end{acronym}

\begin{abstract}
As Large Language Models (LLMs) scale in size and complexity, the consequences of failures during training become increasingly severe. A major challenge arises from Silent Data Corruption (SDC): hardware-induced faults that bypass system-level detection mechanisms. SDC may behave like benign numerical noise, but can also cause harmful gradient corruption that leads to loss spikes, divergence, or stalled progress.

This work provides a controlled study of how intermittent SDC affects LLM pretraining. Using targeted fault injection at the level of GPU matrix-multiply instructions, we characterize the sensitivity of different bit positions, kernel functions, and execution stages. Our analysis shows that locally originating faults can produce impactful corruption, including NaN propagation, short-lived spikes in loss, gradient norm, and attention logits, as well as persistent parameter divergence. Building on the observed corruption signatures, we propose a lightweight detection method that identifies potentially harmful parameter updates.
Experiments on LLaMA models with 60M, 350M, and 1.3B parameters demonstrate that recomputing the most recent training step upon detection can effectively mitigate the impact of these events.
\end{abstract}


\section{Introduction} \label{sec:introduction}

Since the introduction of the Transformer architecture~\cite{vaswani2017attention}, \acfp{LLM} have played a central role in modern artificial intelligence systems.
These models can contain hundreds of billions of parameters and are trained on massive datasets over weeks or months using thousands of hardware accelerators~\cite{touvron2023llama, chowdhery2023palm}. 
As models scale, so too does the complexity and fragility of their training, with frequent reports of instabilities such as loss spikes, exploding or vanishing gradients, divergence, or subtle degradations in performance~\cite{chowdhery2023palm, dehghani2023scaling, zhang2022opt, touvron2021going, molybog2023theory, wortsman2024smallscale, wortsman2023stable, zhai2023stabilizing}. While many of these instabilities stem from model design, training procedures, data quality, or hyperparameter choices, hardware faults remain an underexplored source of disruption~\cite{ma2025understanding}.

A major challenge arises from \acf{SDC}, which silently bypasses system-level detection mechanisms such as \acp{ECC}~\cite{papadimitriou2023silent}. \acp{SDC} can produce incorrect data, such as bit flips, and may remain undetected for extended periods, potentially compromising critical computations and leading to costly debugging efforts~\cite{dixit2021silent}.

One source of \acp{SDC} is transient faults: short-lived errors induced by external factors such as radiation~\cite{ziegler1996terrestrial, mukherjee2005soft, baumann2005radiation}. In terrestrial systems, \ac{SDC} from transient faults are comparatively rare, but with emerging proposals for deploying data centers in space~\cite{starcloud2025}, handling radiation-induced \acp{SDC} may become increasingly important. 

Another root cause is internal hardware faults, such as degradation of aging components or defects introduced during manufacturing~\cite{dixit2021silent, hochschild2021cores, wang2023understanding}. These faults may be permanent or intermittent. Although relatively rare in everyday computing, they are an increasing concern in large-scale systems, where elevated temperatures, workload-induced stress, and the sheer number of devices can significantly increase their likelihood~\cite{dixit2022detecting, wang2023understanding}.

In the context of \ac{LLM} training, vendors such as Google~\cite{team2023gemini}, Meta~\cite{grattafiori2024llama}, ByteDance~\cite{wan2025robust}, and DeepSeek~\cite{zhao2025insights} have highlighted the growing importance of addressing \acp{SDC}. Google’s Gemini models, for instance, encountered \ac{SDC}-related disruptions approximately every one to two weeks~\cite{team2023gemini}, while Meta reported six such incidents during a 54-day training run~\cite{grattafiori2024llama}. DeepSeek emphasizes that \ac{SDC} can silently propagate across many training steps and potentially affect model quality~\cite{zhao2025insights}, while ByteDance notes that a single corrupted gradient can contaminate the global parameter update across all participating workers~\cite{wan2025robust}, where NaN propagation can bring training to an abrupt halt.

A typical approach to handling such faults involves pausing training, locating the source of the fault, and reverting to a previous checkpoint, which introduces considerable overhead. As \acp{LLM} continue to scale and training relies on increasing numbers of hardware accelerators, these interruptions may emerge as a key bottleneck for large-scale training~\cite{wan2025robust}. While such events occur relatively infrequently in practice, their impact can be substantial, as even isolated corruption events may disrupt convergence or degrade model quality.

In this paper, we present a controlled study of \ac{SDC} during \ac{LLM} pretraining through targeted fault injection. Our analysis focuses on single-GPU training to isolate the local computational effects of \ac{SDC}. The implications for distributed training are discussed but not evaluated experimentally. Because training on faulty hardware is rarely feasible and difficult to control, we simulate intermittent faults during computation, representing transient errors that can reoccur under specific environmental or operational conditions such as temperature, workload, or power fluctuations. The contributions of this paper can be summarized as follows:

\begin{itemize}
    \item We conduct a controlled study of \ac{SDC} arising from intermittent faults during \ac{LLM} pretraining based on LLaMA-based models with 60M, 350M, and 1.3B parameters in a controlled single-GPU setting using a custom fault injection framework based on NVBit~\cite{villa2019nvbit}.
    \item We characterize the sensitivity of different bit positions, kernel functions, and execution stages, revealing that locally originating faults can produce impactful corruption of the training process.
    \item Building on the observed corruption signatures, we propose a lightweight detection method that identifies potentially harmful parameter updates associated with loss bumps or NaN values. Upon detection, we show that recomputing the most recent training step can effectively mitigate the impact of these events.
\end{itemize}
The code for this paper is publicly available.\footnote{\url{https://github.com/aaltenbernd/llm-sdc-training}}

\section{Motivation and Related Work}\label{sec:background}

\subsection{Motivation}

\textbf{\acp{ECC} cannot detect all errors that lead to \ac{SDC}.} Memory systems employ \acp{ECC} to enhance resilience against errors caused by hardware faults through the addition of redundant bits to stored data, allowing the system to detect and correct a limited number of bit errors. The most common types, single-bit correction and double-bit detection, can correct one-bit errors and detect, but not fix, two-bit errors~\cite{hamming1950error}. While \acp{ECC} significantly improve system resilience, they are not foolproof. There are several scenarios where \acp{ECC} fail to detect or correct errors, potentially resulting in \ac{SDC}~\cite{meza2015revisiting, papadimitriou2023silent, dixit2021silent}. For instance, \acp{ECC} cannot detect or correct errors that occur during computation nor those arising from unprotected communication paths. Additionally, they are ineffective against errors in memory involving three or more bit flips. As a result, reducing \ac{SDC} requires not only hardware resiliency but also robust fault-tolerant software architectures~\cite{dixit2021silent}.

\textbf{Limitations of \acs{ABFT} for LLM Training.} In contrast to \acp{ECC}, which protect memory, \ac{ABFT} provides protection by incorporating invariants directly into computations to detect and correct errors. \ac{ABFT} for matrix multiplication~\cite{huang1984algorithm} augments the operation with checksum rows and columns, improving reliability at the cost of additional overhead. Moreover, while effective for linear operations, \ac{ABFT} does not readily extend to non-linear operations, as suitable invariants are difficult or impossible to define. As a result, applying \ac{ABFT} comprehensively in large-scale \ac{LLM} training is challenging, since it typically protects only a limited portion of the overall computation while introducing non-negligible overhead relative to the expected rate of \ac{SDC} events.

\subsection{Related Work}

\textbf{\ac{SDC} during inference.} Li et al.~\cite{li2017understanding} were among the first to investigate \ac{SDC} in hardware accelerators, showing that model sensitivity varies by architecture, datatype, and bit position. Agarwal et al.~\cite{agarwal2023resilience} analyzed \ac{SDC} in \acp{LLM} and found that model size and fine-tuning strategy influence vulnerability during inference. Ma et al.~\cite{ma2024dr} monitored deviations in neuron activation distributions as indicators of \ac{SDC} and proposed mitigation through dynamic clipping and probabilistic correction. Sun et al.~\cite{sun2025demystifying} demonstrated that \acp{LLM} are less resilient to random bit flips than previously assumed, revealing significant degradation in output quality across architectures. They further showed that increasing \ac{LLM} scale does not improve resilience.

\textbf{\ac{SDC} during training.} Elsen et al.~\cite{adept-sdc} recommend using deterministic training combined with checksumming of parameters and optimizer states to catch hardware-induced issues across replicas. However, they do not address corruption that arises during the forward or backward pass and subsequently propagates across replicas through gradient aggregation. 

He et al.~\cite{he2023understanding} present a large-scale study of hardware failures in accelerators during model training, showing that intermittent faults can silently corrupt training and degrade convergence. They propose a rollback recovery mechanism based on out-of-bound gradient detection. However, their study does not explicitly focus \ac{SDC} in the context of \ac{LLM} training. Moreover, the proposed detection bound is difficult to justify, as gradient magnitudes during \ac{LLM} training can vary and may reach extreme values. In addition, their method does not distinguish harmful corruption from benign noise, causing any out-of-bound gradient to trigger an alarm and rollback.

Ma et al.~\cite{ma2025understanding} present the first in-depth study of training \acp{LLM} on unhealthy nodes drawn from a production environment. They show that \ac{SDC} can silently corrupt gradients and weights, steering models toward different local minima. During fine-tuning, they observe loss spikes of varying severity. The authors experiment with \ac{ABFT} for matrix multiplication~\cite{huang1984algorithm} to detect \acp{SDC}, but their attempts were mostly unsuccessful, suggesting that the underlying hardware faults in this study occurred outside the matrix multiplication units. Overall, they do not provide actionable detection methods or metrics for identifying \ac{SDC}.

Liang et al.~\cite{liang2025attnchecker} propose ATTNChecker, which uses an \ac{ABFT}-inspired approach to protect attention. Their method can detect and correct extreme errors within attention with modest overhead, outperforming checkpoint-based recovery. However, ATTNChecker protects only a portion of the model, and its additional runtime cost limits its practicality for \ac{LLM} training, while most components remain unprotected.

\section{Methodology} \label{sec:methodology}


\subsection{Experimental Setup} 

\textbf{Models and dataset.} In this study, we adopt the LLaMA-based models~\cite{touvron2023llama} used in~\cite{zhao2024galore, huang2025spam} and experiment with three different model sizes of 60M, 350M, and 1.3B parameters. 
These sizes cover small to moderately large \ac{LLM} scales while remaining tractable for repeated experiments.
We used the English subset of the C4 dataset~\cite{raffel2020exploring} for training.

\textbf{Hyperparameters.} All training runs used a batch size of 512 to ensure stable optimization and practical training throughput\footnote{This effective batch size was achieved via gradient accumulation using micro-batch sizes of 256, 32, and 16 for the different model sizes.}. 
Following Huang et al.~\cite{huang2025spam}, the corresponding maximum learning rates were set to \texttt{1e-3}, \texttt{4e-4}, and \texttt{2e-4}, and we use a learning rate schedule with linear warmup followed by cosine decay. 
The warmup lasted for 1{,}000 steps, and the schedule spanned 100{,}000 steps to reflect a realistic pretraining schedule, even though the full duration was not reached. Training was performed in bfloat16 precision with global gradient norm clipping set to 1.0. We used the AdamW optimizer~\cite{loshchilov2018decoupled} with $\beta_1 = 0.9$, $\beta_2 = 0.999$, and a weight decay of 0.01. These hyperparameters follow common practice for training transformer-based models and were held fixed across all experiments to study the effect of fault injection under a representative and stable training configuration.

\textbf{Deterministic training.} We fixed all relevant sources of randomness using a fixed seed across the involved machine learning libraries. For each experiment, the same seed was used for the fault-injection run and the corresponding baseline for comparison. The dataset was loaded once without streaming so that batches were presented in a deterministic order within each run.

\textbf{Hardware.} All experiments were conducted on a system equipped with a single NVIDIA L40S GPU.

\subsection{Fault Injection}

We use NVBit~\cite{villa2019nvbit}, a dynamic binary instrumentation framework for NVIDIA GPUs, to perform targeted fault injection. NVBit intercepts kernel launches at runtime and instruments SASS instructions, enabling custom instrumentation before or after specific instructions without requiring access to the application source code. Existing NVBit-based fault injection tools primarily support single transient faults or permanent register faults~\cite{tsai2021nvbitfi,guerrero2025effective}. 

To study intermittent computation faults, we implement a custom NVBit instrumentation that targets selected instructions. We insert device-side callbacks before and after the selected instruction. In the pre-instruction callback, we apply a bitmask to the target input register to flip selected bits, and in the post-instruction callback, we restore the original register content to confine the corruption to the computation. We avoid architectural or RTL-level simulation because their runtime overhead~\cite{guerrero2025effective} makes them impractical for \ac{LLM} training workloads, and software-level fault injection because such faults do not propagate realistically through the computation. Although this approach does not fully reproduce real intermittent hardware faults, it preserves realistic instruction-level error propagation with low overhead.

\textbf{Injection control.} Fault injection can be enabled or disabled dynamically during training, allowing control without interrupting execution. The injector provides a set of user-configurable parameters that control where and how faults are introduced, including the \ac{SM}, lane, kernel, instruction, opcode, register, and the bits to flip. Injection behavior is controlled by the training script, which specifies when and where faults are triggered, how frequently they occur, and their duration.

\textbf{Injection scope.} In this study, we restrict our analysis to \ac{HMMA} instructions, as matrix multiplications account for 42.8\%–96.6\% of total execution time in \ac{LLM} workloads~\cite{karami2025understanding}. \ac{HMMA} instructions are executed within \ac{GEMM} kernels throughout the forward and backward pass. During training, \ac{GEMM} kernels appear repeatedly and are invoked for distinct roles, such that faults injected into the same instruction type can propagate differently depending on the kernel context.

\textbf{Fault model.} We focus on \ac{SDC} arising from intermittent faults during computation. Our injection rates exceed those reported for large-scale training systems~\cite{team2023gemini, grattafiori2024llama} and should therefore be interpreted as stress tests that enable systematic analysis within a practical timeframe.

\section{Experimental Findings}\label{sec:findings}

This section characterizes how fault injection affects \ac{LLM} training, including sensitivity to bit positions and kernels (Sections~\ref{subsec:sensitivity_bit_positions} and~\ref{subsec:sensitivity_kernels}),
qualitative differences between faults in the forward and backward pass (Sections~\ref{subsec:forward_pass} and~\ref{subsec:backward_pass}),
spatial effects related to lanes and \acp{SM} (Section~\ref{subsec:spatial_effects}),
and temporal effects arising from fault rate and fault duration (Section~\ref{subsec:temporal_effects}), relative to a fault-free baseline with an evaluation loss of $4.30 \pm 0.006$ obtained from the 60M-parameter model trained for 1{,}000 steps across ten initialization seeds.

\subsection{Sensitivity of Bit Positions}\label{subsec:sensitivity_bit_positions}

In the first experiment, the objective was to identify relevant bit positions for subsequent analyses. Based on the best performing seed, we conducted a fault injection campaign spanning 1,000 training steps. All experiments were fixed to a single \ac{SM} and lane, and we varied the bitmask over the set $\{2^i \mid i \in \mathbb{N}, 0 \le i \le 31\}$, which correspond to bit positions 0–31. We selected one out of five \ac{HMMA} instructions in each kernel function and injected faults in one out of every ten training iterations, both chosen at random. Fault injection was applied to the first input register.

\textbf{Only exponent bits exhibited measurable impact.} For bit positions {10, 11, 26, 27}, we observed increases in evaluation loss that exceeded the baseline variation. For bit positions {12, 13, 14, 28, 29, 30}, the evaluation produced NaN values. Mantissa bits and the sign bit did not show noticeable differences, and therefore only the exponent bits exhibited measurable effects. The effects were nearly symmetric across the input register, which reflects the structure of \ac{HMMA} instructions, where each register packs two bfloat16 values.

\subsection{Sensitivity of Kernels}\label{subsec:sensitivity_kernels}

In the second experiment, we therefore restrict the analysis to the first input value in the first input register, specifically its seven highest exponent bits and its sign bit, the latter included because of its unique characteristics. We conducted again a fault injection campaign for the first 1,000 training steps using the best performing seed. We fixed all experiments to a single \ac{SM} and lane, and varied the bitmask in the predetermined set $\{2^i \mid i \in \mathbb{N},\, 8 \le i \le 15\}$. In addition, we limited each fault injection run to a single kernel function. For the 60M model we identified five \ac{GEMM} kernels in the forward pass and ten \ac{GEMM} kernels in the backward pass. In the following, we denote these as $\text{FP}_i$ and $\text{BP}_i$, referring to the i-th \ac{GEMM} kernel in the forward or backward pass, respectively. In this experiment, we injected faults into each \ac{HMMA} instruction in these kernels and maintained the schedule of injecting faults randomly in one out of ten training iterations. 

We evaluate the effect of these faults using four metrics: the evaluation loss, the parameter difference relative to the corresponding baseline, the gradient norm before clipping, and the maximum attention logits. The parameter difference is defined as the L2 norm of the absolute pairwise parameter differences relative to the corresponding baseline~\cite{ma2025understanding}. Because we fixed the random seeds of the machine learning libraries in our training environment, the parameter difference remains zero if no fault is injected. The attention logits are the values produced by the attention mechanism before the softmax and can approach one hot attention vectors when extremely large. Maximum attention logits have been used as a signal of training instability in previous work~\cite{dehghani2023scaling, wortsman2024smallscale}.

\begin{figure}[t]
\centering
\includegraphics[width=\columnwidth]{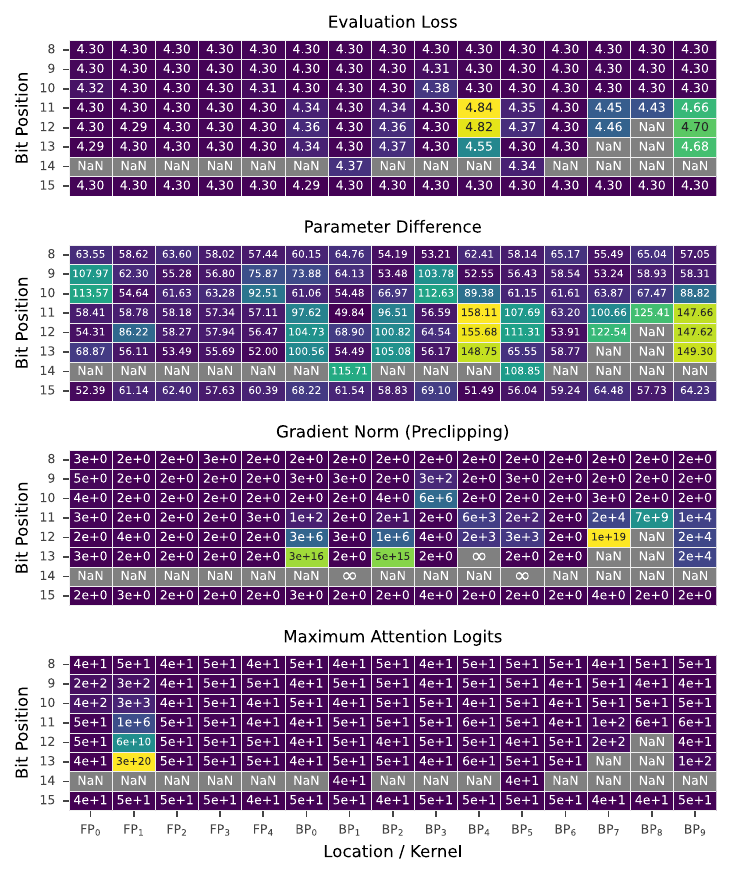}
\caption{Evaluation loss, parameter difference, gradient norm before clipping, and maximum attention logits for different bit positions and kernels ($\text{FP}_i$ and $\text{BP}_i$, referring to the i-th \ac{GEMM} kernel in the forward or backward pass, respectively). Evaluation loss and parameter difference represent the value observed after the training span. The gradient norm and the maximum attention logits represent the maximum values observed over the training span.}
\label{fig:bitpos_vs_kernel}
\end{figure}

Figure~\ref{fig:bitpos_vs_kernel} shows the evaluation loss, the parameter difference, the gradient norm before clipping, and the maximum attention logits for each combination of bit position and kernel function in the forward and backward pass. Both the gradient norm and the maximum attention logits in the figure represent the maximum of all observed values in the training span. For the parameter difference and the evaluation loss, the value was determined after the training span. 

\textbf{Sensitivity of bit positions varies across kernels.} For most bit positions the evaluation loss remains close to the baseline, which indicates that many injected faults do not introduce substantial disruptions during training. Certain bit positions, however, lead to clear deviations where the evaluation loss increases above the baseline variation. This indicates localized sensitivity to faults in specific kernels. With regard to NaN propagation, almost all kernels produce NaN values when injecting the \ac{MSB}, and some functions show sensitivity even for lower bit positions. Despite NaN values appearing in the forward pass when injecting the \ac{MSB}, measurable effects on the evaluation loss arise almost exclusively in the backward pass.

\textbf{Parameter shifts occur in all kernels.} Although these events do not always lead to an increase in evaluation loss, \acp{SDC} can nevertheless lead to unwanted parameter changes, where all runs deviate from their corresponding fault-free baselines. Runs with increased evaluation loss also display relatively large parameter differences. However, larger parameter differences do not necessarily result in larger evaluation loss. This asymmetry is expected, as we assume that the impact depends strongly on which parameters are affected. 

\textbf{Gradient norm exhibits high values.} The gradient norm before clipping remains stable at approximately $2\mathrm{e}{+}0$ in most runs. Elevated values, ranging from $1\mathrm{e}{+}2$ to $1\mathrm{e}{+}19$, appear only when faults are injected into kernels of the backward pass. In some runs, including injections of the \ac{MSB} in $\text{BP}_1$ and $\text{BP}_5$ and the 13th bit in $\text{BP}_4$, the gradient norm before clipping even jumps to infinity. The absence of such spikes in the forward pass indicates that corruptions introduced during the forward pass have only limited influence on the subsequent learning dynamics, whereas faults in the backward pass directly lead to gradient corruption and thus have a much stronger impact on the training.

\textbf{Extreme attention logits.} The maximum attention logits behave in a similar manner. Most runs fall within the expected range, but some injections produce extremely large values. Notably, these outliers appear only in a single kernel of the forward pass, $\text{FP}_1$, which appears to be responsible for computing the attention logits before softmax. Bit positions 10 to 13 in this kernel result in values between $3\mathrm{e}{+}3$ and $3\mathrm{e}{+}20$. Such extreme attention logits are likely to produce near one hot attention vectors and may momentarily disrupt the attention mechanism.

\subsection{Transient Effects in the Forward Pass}\label{subsec:forward_pass}

While the forward pass is generally robust to faults, it can still exhibit transient effects. This section examines spikes in the loss and maximum attention logits.

\textbf{Transient spikes appear in the loss.} Although the forward pass exhibits limited sensitivity to the fault injections, the training loss remains a useful indicator of \ac{SDC}. For $\text{FP}_0$ and $\text{FP}_4$ we occasionally observe small spikes in the training loss for bit positions 9 and 10. However, these spikes return immediately to the previous level. Such spikes may also arise from training dynamics and therefore cannot be distinguished from normal training behavior. Nevertheless, they may indicate corruption events, which must be supported by stronger signals.

\textbf{Attention logit spikes reveal corruption.} In addition, $\text{FP}_1$ shows abrupt jumps in the maximum attention logits. Under normal conditions, these logits evolve smoothly across training iterations rather than exhibiting sudden and large spikes. Even unusually large updates to the attention parameters would not produce such changes with such high magnitude, as the attention logits depend linearly on the query and key projections. A sharp spike in the maximum attention logits is therefore a strong indicator of corruption. More generally, monitoring sudden changes in other intermediate results may also provide useful information for identifying corruption events.

\subsection{Severe Effects in the Backward Pass}\label{subsec:backward_pass}

Compared to the forward pass, the backward pass is far more sensitive to injected faults. As in the forward pass, sudden spikes in the gradient norm can appear before clipping. Such spikes may also arise from normal training dynamics, so they cannot be used alone to identify corruption. However, faults introduced during the backward pass tend to have much more severe consequences because corrupted gradients directly influence parameter updates.

\begin{figure}[t]
\centering
\includegraphics[width=\columnwidth]{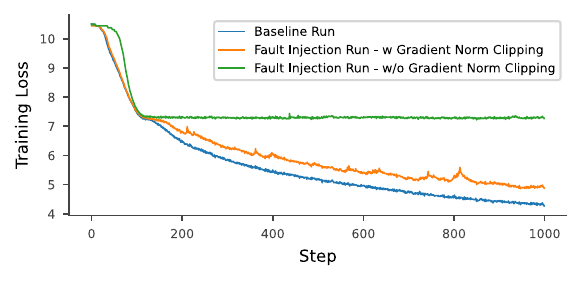}
\caption{Training loss for fault-injection runs with and without gradient norm clipping, and for the corresponding baseline run (blue). With gradient norm clipping enabled (orange), injected faults create repeated loss spikes but training continues, illustrating that gradient norm clipping mitigates most harmful spikes. Without gradient norm clipping (green), a single corrupted gradient causes the optimizer’s second moment to become infinity, freezing parameter updates and stalling training.}
\label{fig:loss_gc}
\end{figure}

\textbf{Clipping mitigates most severe effects.}
If gradient norm clipping is disabled, training may halt completely due to how the second moment is updated in adaptive optimizers. Because the second moment scales with the squared gradient, extremely large gradients can cause it to diverge to infinity, which in turn drives the adaptive update to zero. When this occurs, the optimizer stops updating the affected parameters. As shown in Figure~\ref{fig:loss_gc}, when gradient norm clipping is disabled for $\text{FP}_9$ at bit position~13 and the fault injection rate is kept unchanged (green curve), training stalls entirely.

\textbf{Gradient corruption persists even with clipping.}
Even with gradient norm clipping enabled, we observed loss bumps resulting from faults injected into the backward pass. This is illustrated in Figure~\ref{fig:loss_gc} for $\text{FP}_9$ and bit position 13 (orange curve). These faults caused gradient corruption, most notably in the form of gradient spikes. The relationship between gradient spikes and loss bumps has already been discussed in~\cite{huang2025spam}, where the authors classify every loss bump as harmful to the overall training progress. Adaptive optimizers such as Adam~\cite{kingma2014adam} can exacerbate this effect because accumulated momentum amplifies the influence of corrupted gradients. To mitigate this issue, the authors propose momentum resets for parameters whose gradients exhibit spike behavior. It remains an open question whether momentum reset can serve as a proactive mitigation for \ac{SDC}, as \ac{SDC} can cause widespread gradient corruption and may trigger numerous momentum resets.

\textbf{Infinite gradient norm breaks the clipping.}
Gradient norm clipping typically mitigates the most severe of these events. However, we identified an unintended interaction: when the gradient norm before clipping becomes infinite, the clipped gradient norm collapses to zero. As a result, all gradients are zeroed and the corresponding mini-batch is effectively ignored. This behavior follows directly from the definition of gradient norm clipping. Formally, let \(\lVert G_t \rVert_{\text{pre}}\) denote the global gradient norm before clipping at iteration \(t\). Gradient norm clipping defines the clipped gradient as
\begin{equation}
\tilde{g}_t =
\begin{cases}
g_t & \text{if } \lVert G_t \rVert_{\text{pre}} \le \tau, \\[6pt]
\dfrac{\tau}{\lVert G_t \rVert_{\text{pre}}} g_t & \text{otherwise},
\end{cases}
\end{equation}
where $\tau$ is the threshold for the maximum allowed gradient norm. If $\lVert G_t \rVert_{\text{pre}}=\infty$, the scaling factor becomes $\tau/\infty=0$, which forces $\tilde{g}_t=0$. 

\textbf{Distributed training amplifies this risk.}
In data-parallel training, where gradients are aggregated across replicas, the interaction with gradient norm clipping may become particularly severe. If one replica produces extremely large gradient values, these values propagate into the gradient aggregation process. As a result, the global gradient norm can become infinite after aggregation. This means that the work of all replicas, including potentially thousands of devices in large-scale training, is effectively discarded for that iteration. 

\textbf{Momentum-only updates.}
Gradient corruption can have consequences for the optimization process, ranging from mild deviations in flat regions of the loss surface to significant disruptions in regions that are highly sensitive to parameter changes. When gradient clipping collapses all gradients to zero, this effect propagates globally, since the optimizer receives no gradient signal for that iteration and therefore performs its update using only accumulated momentum. Such momentum-only updates apply to all parameters and can cause the optimizer to continue moving in a direction that may be outdated. It remains an open question whether this interaction with gradient norm clipping poses an even greater risk to training stability in highly sensitive regions of the loss surface.

\subsection{Spatial Effects: Lanes and SMs}\label{subsec:spatial_effects}

To isolate the effect of fault location within the GPU, we conducted an additional experiment in which we fixed the injection site to $\text{BP}_9$ and the bit position to 13. We examined the impact of varying both the lane and the \ac{SM}, using the same injection schedule as in previous experiments.

\textbf{No sensitivity to lane selection.} We first examined the impact of varying the lane while keeping the \ac{SM} fixed. Faults were injected across all 32 lanes. Across all runs, the evaluation loss remained stable at $4.60 \pm 0.08$. These results indicate that the sensitivity of $\text{BP}_9$ to this bit position is largely independent of the specific lane, suggesting that similar fault effects arise across all lanes and that intra-warp variation does not lead to systematically different fault propagation behaviors.

\textbf{No sensitivity to \ac{SM} selection.} In the second part of the experiment, we varied the \ac{SM} while fixing the lane. To sample the device uniformly, we evaluated every fourth \ac{SM}. Similar to the lane experiment, the evaluation loss remained within a narrow range of $4.63 \pm 0.07$. This consistency across \acp{SM} suggests that similar corruption effects occur regardless of which \ac{SM} executes the kernel, at least for $\text{BP}_9$ and bit position~13.

\subsection{Temporal Effects: Fault Rate and Duration}\label{subsec:temporal_effects}

In this experiment, we examined how the temporal characteristics of a fault influence its impact by varying both the injection rate and the duration for which a fault remains active.

\textbf{Higher fault rates increase impact.}  
To assess the effect of the injection rate, we repeated the fault injection for $\text{FP}_9$ and bit position~13 using different rates: every step, one out of ten steps, one out of one hundred steps, and one out of one thousand steps. Using all ten seeds, the resulting evaluation losses for these rates were $7.38 \pm 0.073$, $4.77 \pm 0.046$, $4.33 \pm 0.018$, and $4.30 \pm 0.003$, respectively. These results show a clear dependence on the fault rate. High rates lead to substantially larger evaluation losses, while lower rates reduce the impact toward the level of fault-free execution. At the highest injection rate, faults affect every parameter update. As a result, the optimizer cannot recover between steps and the training dynamics saturate, leading to a plateau in the observed degradation. Reducing the injection rate allows recovery between injections, explaining the large drop when moving from faults at every step to every ten steps.

\textbf{Longer fault durations increase deviation.}  
To assess the effect of duration, we injected a fault at step 500 and kept it active for 1, 3, 5, 7, or 9 consecutive steps. Again using all ten seeds, the resulting evaluation losses were $4.30 \pm 0.005$, $4.31 \pm 0.006$, $4.32 \pm 0.007$, $4.33 \pm 0.008$, and $4.35 \pm 0.009$ for the respective fault durations. These results show that longer fault durations cause progressively larger deviations. In particular, both the magnitude of the resulting loss bump and the time required for recovery increase with fault duration.

\section{Harmful Parameter Updates}\label{sec:harmful_parameter_updates}

The preceding analysis shows that injected faults can cause disruptions in gradients, yet the long-term impact of these disruptions depends on how the optimizer transforms gradients into parameter updates. Ultimately, it is the optimizer, not the raw gradients, that determines whether a corruption event meaningfully alters the model’s trajectory. To study this connection, we analyze the parameter update rule of AdamW~\cite{loshchilov2018decoupled}:
\begin{align}
m_t &= \beta_1 m_{t-1} + (1-\beta_1)\, g_t, \\[4pt]
v_t &= \beta_2 v_{t-1} + (1-\beta_2)\, g_t^2, \\[8pt]
\hat{m}_t &= \frac{m_t}{1-\beta_1^{t}}, \\[4pt]
\hat{v}_t &= \frac{v_t}{1-\beta_2^{t}}, \\[8pt]
\theta_{t+1}
&= \theta_t
    - \eta \left(
        \frac{\hat{m}_t}{\sqrt{\hat{v}_t} + \epsilon}
        + \lambda\,\theta_t
      \right).
\end{align}

Here, $m_t$ and $v_t$ denote exponential moving averages of the first and second moments of the gradient, computed using decay rates $\beta_1$ and $\beta_2$, respectively. The bias-corrected quantities $\hat{m}_t$ and $\hat{v}_t$ converge to their uncorrected counterparts after a few iterations. The learning rate is denoted by $\eta$, $\lambda$ controls weight decay, and $\epsilon$ is a small constant ensuring numerical stability. The parameter at iteration $t$ is $\theta_t$, and $\theta_{t+1}$ is the updated parameter.

Since the weight decay is independent of the gradient, its contribution is unaffected by gradient corruption. We therefore focus on the adaptive component of the update, which (omitting bias correction for clarity, which is valid after a few iterations) is given for each parameter $p$ by
\begin{equation}
    U_{t,p} = \eta\, \frac{m_{t,p}}{\sqrt{v_{t,p}} + \epsilon}.
\end{equation}

To quantify the magnitude of these updates, we track the maximum \ac{RMS} update across all parameter groups $\mathcal{P}$,
\begin{equation}
    R_t
    =
    \max_{P \in \mathcal{P}}
    \left(
        \frac{1}{|P|}
        \sum_{p \in P} U_{t,p}^2
    \right)^{1/2},
\end{equation}
which measures the average update magnitude within each group, gives more weight to larger updates, and identifies the group with the largest such update magnitude. The quantity $R_t$ thus serves as a useful indicator of unusually large updates to the parameters. 

\begin{figure}[t]
\centering
\includegraphics[width=\columnwidth]{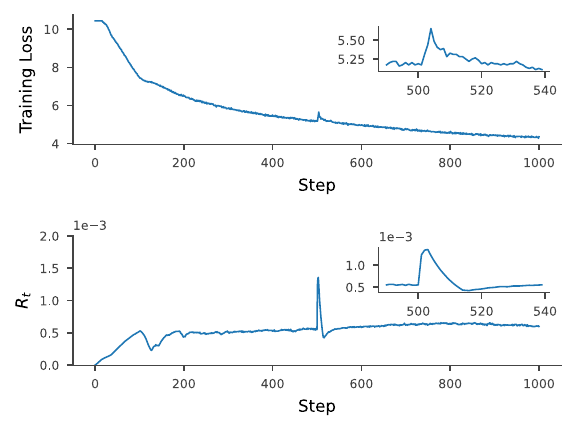}
\caption{Effect of a single fault injection with a fault length of three on the training loss and the parameter update magnitude $R_t$. The injected fault triggers a sharp spike in $R_t$, after which the training loss exhibits a bump before gradually stabilizing at a slightly elevated level. This example illustrates how gradient corruption propagates through the optimizer and leads to harmful parameter updates.}
\label{fig:rms}
\end{figure}

In the experiments where we varied the fault duration in Subsection~\ref{subsec:temporal_effects}, we observed characteristic changes in $R_t$ following fault injections. In all runs, $R_t$ exhibited a sudden spike immediately after the fault was introduced. Subsequently, the training loss increased gradually, forming a distinct loss bump before stabilizing again at a slightly higher level than before the disturbance. Both the height of the loss bump and the recovery time grew with increasing fault length. Likewise, the fault length determined how long $R_t$ required to decay back to its stable range, with longer fault durations leading to more prolonged deviations. In Figure~\ref{fig:rms}, we show an example for a single injection and a fault length of three. 

These experiments indicate that sudden spikes in $R_t$ caused by gradient corruption directly lead to loss bumps. Since loss bumps are considered harmful to the training progress~\cite{huang2025spam}, detecting such spikes in $R_t$ is essential for identifying \textit{harmful parameter updates} induced by \ac{SDC}.

\section{Detection}\label{sec:detection}

As previously shown, \acp{SDC} can lead either to benign noise or to harmful parameter updates. To detect the latter, we measure the absolute change in $R_t$ between consecutive steps,
\begin{equation}
\Delta R_t = \bigl| R_t - R_{t-1} \bigr|.
\end{equation}

We compute the empirical mean $\mu$ of $\Delta R_t$ observed during the warm-up phase and use this as a baseline for comparison:
\begin{equation}
\mu = \frac{1}{T} \sum_{i=1}^{T} \Delta R_i,
\end{equation}
where $T = \min(t, t_w)$ and $t_w$ denotes the warm-up length.

To detect deviations from this baseline, we can flag a jump in $R_t$ as anomalous whenever
\begin{equation}
\Delta R_t > \frac{\mu}{\alpha}, \label{eq:detection1}
\end{equation}
where $\alpha \in (0,1]$ is a configurable sensitivity parameter. Larger values of $\alpha$ make the detector more sensitive to deviations, while smaller values make it more conservative.

The choice of $\alpha$ inherently presents a tradeoff between sensitivity and specificity. Besides jumps in $R_t$, we also observe jumps in the global gradient norm when injecting faults. To better navigate this tradeoff, we account for positive jumps in the gradient norm by defining
\begin{equation}
\Delta G_t = \max\left(0, \lVert G_t \rVert_{\text{pre}} - \lVert G_{t-1} \rVert_{\text{pre}} \right),
\end{equation}
where $\lVert G_t \rVert_{\text{pre}}$ denotes the gradient norm before clipping.

Since gradient norm jumps provide useful context but are not sufficient on their own to indicate harmful parameter updates, we use them as an auxiliary signal in the detection and refine Equation~(\ref{eq:detection1}):
\begin{equation}
    \Delta R_t > \frac{\mu}{\alpha \sqrt{\Delta G_t}}
\end{equation}

Incorporating $\Delta G_t$ modulates the detector’s sensitivity by adjusting how strongly changes in $R_t$ contribute to anomaly decisions. When $\Delta G_t$ is small, the detector becomes less responsive to avoid false positives, while larger $\Delta G_t$ increases responsiveness. The division by $\sqrt{\Delta G_t}$ reflects that $R_t$ is an RMS quantity that scales with the square root of the underlying parameter update magnitudes.

Additionally, we flag the current training step as anomalous if $\lVert G_t \rVert_{\text{pre}}$ is not finite. As shown in Section~\ref{sec:findings}, a gradient norm of infinity causes the clipped gradient norm to collapse to zero, and NaN values halt training. Both indicate failure states and are therefore flagged as anomalous.

While \ac{SDC} can be the cause of harmful parameter updates, such updates may also originate from other sources. In practical training environments, the proposed detection mechanism should therefore be integrated with system-level metrics such as temperature, utilization, and power consumption in order to provide a broader view. These system-level metrics could enable context-aware modulation of the sensitivity parameter $\alpha$, allowing the detector to adapt its sensitivity to changing hardware conditions.

When a harmful parameter update is detected, we recompute the current training step to verify whether the detection was valid and to mitigate the effect. Under healthy training conditions, this recomputation should produce the same value of $\Delta R_t$, since identical input data and an unchanged model state yield the same output. If the original computation was corrupted, the recomputed $\Delta R_t$ will most likely differ because hardware scheduling is nondeterministic.

The proposed detection mechanism is compatible with distributed training. The detection metric can be computed locally from gradients and optimizer statistics on each worker, allowing corrupted steps to be detected and recomputed before gradients propagate and enabling localization of the faulty device. Alternatively, the metric can be evaluated after gradient aggregation, yielding identical detection decisions across workers and avoiding independent false positives across devices that could otherwise lead to frequent recomputations at large scale.

\begin{figure*}[t]
\centering
\includegraphics[width=\textwidth]{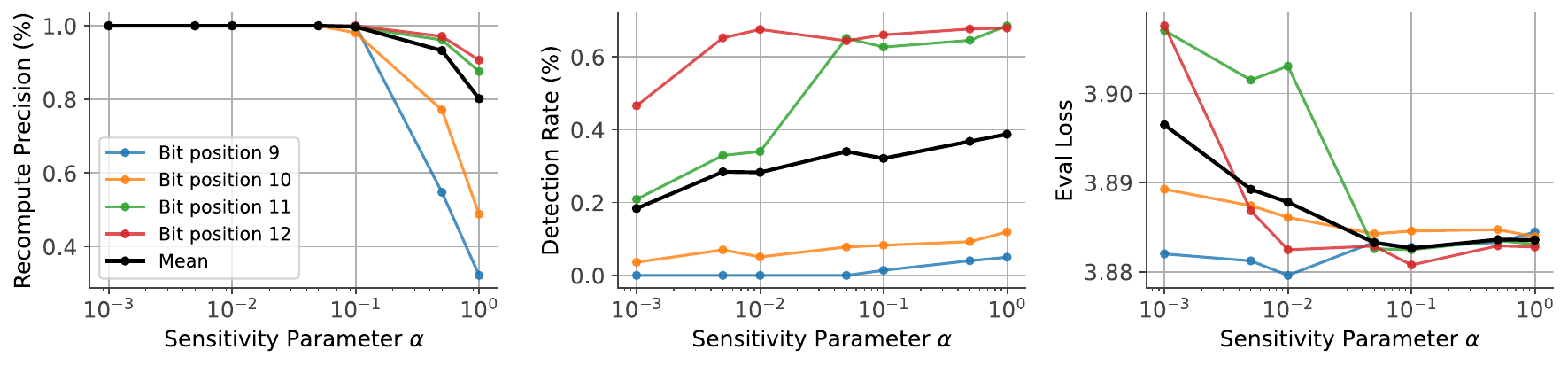}
\caption{
Effect of the detector sensitivity parameter $\alpha$ on recomputation behavior over 2{,}000 training steps for the 60M-parameter model. Results are shown per bit position and averaged across all bit positions (in black).
\textbf{Left:} Recompute precision, indicating how often recomputation corresponds to true corruption. Increasing $\alpha$ raises the false alarm rate, leading to unnecessary recomputation.
\textbf{Center:} Detection rate, showing increased sensitivity as $\alpha$ increases.
\textbf{Right:} Evaluation loss, which exhibits an increase for smaller $\alpha$ relative to the baseline variation ($3.881 \pm 0.002$, after 2,000 steps).
}
\label{fig:confusion}
\end{figure*}

\section{Evaluation}\label{sec:evaluation}

This section evaluates the proposed detection mechanism with respect to its ability to distinguish harmful parameter updates from benign noise and to limit training progress loss through selective recomputation.

\subsection{\texorpdfstring{Impact of the Sensitivity Parameter $\alpha$}{Impact of the Sensitivity Parameter alpha}}

We study the effect of the sensitivity parameter $\alpha$ on detection behavior using the 60M model trained for 2{,}000 steps with three seeds, evaluating recompute precision, detection rate, and evaluation loss. 

We define recompute precision as the fraction of detected anomalies for which recomputation is justified, and thus as an indirect measure of the false alarm rate. The detection rate captures the fraction of injected faults that are successfully identified and thus the fraction that triggers recomputation. While we report the detection rate, we do not report conventional metrics such as recall or F1-score, as many \acp{SDC} are masked during training and do not result in observable degradation. Consequently, such metrics provide limited insight into the effectiveness of detection and recomputation.

We evaluate discrete values of $\alpha \in [10^{-3}, 1]$ on a logarithmic scale. During training, faults are injected once every ten steps on average, with the target kernel selected at random. This analysis is restricted to kernels in the backward pass and to bit positions 9 to 12. Higher bit positions are more likely to produce NaN values. While easier to detect, they are less informative for evaluating detector sensitivity. Conversely, lower bit positions are less likely to have a measurable impact. Together, these bit positions provide a representative range for assessing the effect of $\alpha$. 

Figure~\ref{fig:confusion} summarizes the results. Larger values of $\alpha$ increase detector sensitivity, improving detection rates but raising the false alarm rate and resulting in unnecessary recomputations. Smaller values reduce false positives but miss a larger fraction of injected faults. The evaluation loss increases slightly as $\alpha$ decreases, indicating that overly conservative recomputation fails to fully mitigate harmful updates. These results highlight the trade-off between detector sensitivity and recomputation overhead. $\alpha = 0.05$ provides a favorable balance between detector sensitivity and recomputation overhead, preventing divergence while avoiding unnecessary recomputations.

\subsection{Fault Injection and Recomputation Across Model Scales}

We next evaluate the effectiveness of detection and recomputation across model scales. Experiments were repeated across multiple initialization seeds: twelve runs for the 60M model, six for the 350M model, and three for the 1.3B model, reflecting the increased computational cost at larger scales. All models were trained for 10{,}000 steps, and evaluation loss was recorded every 1{,}000 steps. Final evaluation losses are reported in Table~\ref{tab:eval_loss_runtime_overhead}. For all experiments in this section, we set $\alpha=0.05$, for which no false positives were observed during baseline training.

While our experiments focus on LLaMA-based architectures, the evaluated models span more than an order of magnitude in parameter count, allowing us to study the behavior of the proposed method across different training scales. Because the detection mechanism operates on optimizer statistics rather than model-specific structure, it is expected to generalize to other transformer-based architectures.

We repeated all experiments with fault injection enabled. In these runs, we restricted the fault configuration to $\text{BP}_4$ and $\text{BP}_9$, as these settings produce the most pronounced impact. Faults were injected once every 100 training steps on average, with a randomly sampled duration between 1 and 5 steps. For each fault, the target kernel was selected at random. To ensure consistent fault targeting across model sizes, we had to map kernel names, as the 350M and 1.3B models expose a different set of kernel launches than the 60M model.

The bit position was fixed within each run to match the behavior of intermittent faults. For the 60M and the 350M model, faults were injected at bit position 11 for half of the seeds and at bit position 12 for the remaining seeds. These bit positions yield substantial numerical impact while keeping the probability of NaN values low. Avoiding NaNs is important both for efficient use of computational resources and for maintaining interpretability of the results. For the 1.3B model, we used two seeds with bit position 12 and one seed with bit position 11. 

Finally, we repeated all experiments with fault injection enabled while recomputing the most recent training step whenever an anomaly was detected. For simplicity, fault injection was not reactivated during recomputation, although this scenario is possible in practice.

Across all model sizes, fault injection led to a substantial increase in evaluation loss after 10{,}000 steps, as shown in Table~\ref{tab:eval_loss_runtime_overhead}. In contrast, recomputation effectively mitigated the impact of injected faults, and the evaluation loss remained close to the fault-free baseline.

\begin{table}[t]
    \centering
    \caption{Evaluation Loss and Runtime Overhead}
    \label{tab:eval_loss_runtime_overhead}
    \resizebox{\columnwidth}{!}{
        \begin{tabular}{l c c c}
            \hline \noalign{\vskip 2pt}
            \textbf{Model Size} & 60\textbf{M} & 350\textbf{M} & 1.3\textbf{B} \\ 
                   & (12 seeds)   & (6 seeds)     & (3 seeds)     \\ 
            \noalign{\vskip 2pt} \hline \noalign{\vskip 2pt}
            Baseline Runs & $3.50 \pm 0.002$ & $3.25 \pm 0.001$ & $3.26 \pm 0.001$ \\
            Runs with Fault Injection & $3.61 \pm 0.037$ & $3.37 \pm 0.024$ & $3.34 \pm 0.005$ \\
            Runs with Recomputation & $3.50 \pm 0.002$ & $3.25 \pm 0.001$ & $3.26 \pm 0.001$ \\ \noalign{\vskip 2pt} \hline \noalign{\vskip 2pt}
            
            Runtime w/o Detection & 0.98 s/it & 3.93 s/it & 11.61 s/it \\
            Runtime w Detection & 0.99 s/it & 3.95 s/it & 11.68 s/it \\ \noalign{\vskip 2pt} \hline
        \end{tabular}
    }
\end{table}

\subsection{Runtime Overhead}

Table~\ref{tab:eval_loss_runtime_overhead} also shows the runtime overhead when detection is enabled, accounting for roughly $1\%$ of the training throughput. This overhead must be weighed against the training progress lost under undetected fault injection. Comparing the evaluation loss of fault-free baselines at earlier steps shows that the final loss under fault injection corresponds to the loss observed approximately 3{,}000 to 4{,}000 steps earlier in baseline runs, indicating a substantial and persistent loss of training progress. The fault injection rates used in our experiments are intended as stress tests rather than to reflect realistic fault frequencies in production systems. Nevertheless, these experiments illustrate that even isolated corruption events can lead to lasting deviations in the training trajectory. On average, each injected fault resulted in a loss of approximately 30 to 40 training steps.

Importantly, the cost of detection and the cost of faults scale differently. Detection introduces a fixed per-iteration overhead that is predictable and independent of fault frequency. In contrast, the impact of undetected faults can be amplified by the training process itself. Under data-parallel training, a single corrupted gradient can propagate through gradient aggregation and affect all replicas, while corruption of optimizer state may persist across multiple subsequent steps. As model size and system parallelism increase, the resulting loss of training progress may therefore grow disproportionately.

In practice, faults are commonly handled by reverting training to an earlier checkpoint, which can discard substantial training progress. In contrast, recomputing the most recent training step upon detection instead restores the fault-free training trajectory at the cost of a bounded overhead. This trade-off makes detection and recomputation a practical and effective mechanism for mitigating \ac{SDC} during LLM training. In the presence of frequent faults, however, the overhead of repeated recomputation may become prohibitive, making device replacement unavoidable.

\subsection{Limitations}

Our study uses controlled fault-injection experiments in a single-GPU setting to isolate how computation-level corruption propagates through training. However, it does not capture all effects that may arise in large-scale distributed environments, where gradient aggregation and system-level fault handling may influence how corruption propagates across replicas.

While the proposed detection metric can be computed either locally on each worker or after gradient aggregation, we do not experimentally evaluate these scenarios in a distributed setting. Studying how faults propagate across replicas and how recomputation interacts in large-scale systems remains an important direction for future work.

Furthermore, although our experiments span models from 60M to 1.3B parameters, evaluating even larger models would strengthen the empirical validation of the proposed method. In addition, our experiments focus solely on LLaMA-based models trained with the AdamW~\cite{loshchilov2018decoupled} optimizer. Evaluating other architectures and optimization methods would further validate the generality of the approach.

Finally, our fault model focuses on \ac{GEMM} kernels. While matrix multiplications dominate \ac{LLM} workloads, faults in other operations or hardware components may also influence training behavior.

\section{Conclusion}\label{sec:conclusion}

In this paper, we presented a controlled study of \ac{SDC} during LLM pretraining and demonstrated that intermittent hardware faults can induce harmful parameter updates that lead to loss bumps and substantial degradation of training progress. 
Based on these observations, we proposed a lightweight detection mechanism that identifies harmful parameter updates and demonstrated experimentally that recomputing corrupted steps can effectively mitigate their impact. 
Our results highlight \ac{SDC} as a significant reliability challenge in large-scale \ac{LLM} training, and we hope to motivate the development of practical detection, localization, and mitigation techniques.

A promising direction for future work is the integration of system-level metrics, such as temperature, utilization, and power consumption, with model-level metrics to enable more accurate detection and attribution of faults. In distributed training, further research is needed to characterize how failures propagate through gradient aggregation and to design mitigation strategies that prevent system-wide degradation. Future work should also explore proactive mitigation methods, such as momentum resets~\cite{huang2025spam}, which may reduce the need for recomputation and help maintain stable training progress. Finally, incorporating low-cost invariant checks could provide lightweight, continuous validation of system and model behavior, enabling earlier and more precise fault localization.